\documentclass[11pt,a4paper]{scrartcl}
\usepackage{booktabs}

\usepackage{color}
\usepackage{latexsym}              
\usepackage{amsmath}
\usepackage{amssymb}               
\usepackage{amsfonts}              
\usepackage{amsthm}                
\usepackage{mathtools}
\usepackage{multirow}
\usepackage{tikz}                  
\usetikzlibrary{arrows,positioning,shapes}
\usepackage{tcolorbox}
\usepackage{pifont}

\usepackage[english]{babel} 

\RequirePackage[%
  pdfstartview=FitH,%
  breaklinks=true,%
  bookmarks=true,%
  colorlinks=true,%
  linkcolor= blue,
  anchorcolor=blue,%
  citecolor=orange,
  filecolor=blue,%
  menucolor=blue,%
  urlcolor=blue%
  ]{hyperref}
  
  \AtBeginDocument{%
  \hypersetup{%
    pdfauthor={M Perez-Ortiz},%
    pdftitle={Title - compilation : \today}%
    }
    }

\usepackage{fancyhdr}
\newcommand{\cmark}{\ding{51}}%
\newcommand{\xmark}{\ding{55}}%

\usepackage[mathcal]{eucal}
\usepackage{cleveref}
\crefname{assumption}{Assumption}{Assumptions}
\crefname{equation}{Eq.}{Eqs.}
\crefname{figure}{Fig.}{Figs.}
\crefname{table}{Table}{Tables}
\crefname{section}{Sec.}{Secs.}
\crefname{theorem}{Thm.}{Thms.}
\crefname{lemma}{Lemma}{Lemmas}
\crefname{corollary}{Cor.}{Cors.}
\crefname{example}{Example}{Examples}
\crefname{appendix}{Appendix}{Appendixes}
\crefname{remark}{Remark}{Remark}

\makeatother

\theoremstyle{plain}  

\def\ddefloop#1{\ifx\ddefloop#1\else\ddef{#1}\expandafter\ddefloop\fi}

\def\ddef#1{\expandafter\def\csname c#1\endcsname{\ensuremath{\mathcal{#1}}}}
\ddefloop ABCDEFGHIJKLMNOPQRSTUVWXYZ\ddefloop

\newcommand{\R}{\mathbb{R}}
\newcommand{\hL}{\hat{L}}
\newcommand{\ce}{\text{x-e}}
\newcommand{\KL}{\operatorname{KL}} 

\begin{document}
\title{Learning PAC-Bayes Priors\\for Probabilistic Neural Networks}

\author{
\small{
    Mar\'ia P\'erez-Ortiz$^*$, 
    Omar Rivasplata$^*$, 
    Benjamin Guedj$^{*\dagger}$, 
    Matthew Gleeson$^*$, 
}
\\ 
\small{
    Jingyu Zhang$^*$, 
    John Shawe-Taylor$^*$, 
    Miroslaw Bober$^\S$ and 
    Josef Kittler$^\S$
}
\\[2ex] 
\small{
    $^*$Centre for AI and Dept. of Computer Science, University College London (UK)
}
\\ \newline
\small{
$^\dagger$Inria, Lille Nord-Europe research centre and Inria London Programme (France)
}
\\ \newline
\small{
$^\S$Center for Vision, Speech and Signal Processing (CVSSP), University of Surrey (UK)
}
}

\date{ }

\maketitle

\begin{abstract}
Recent works have investigated deep learning models trained by optimising PAC-Bayes bounds, with priors that are learnt on subsets of the data. This combination has been shown to lead not only to accurate classifiers, but also to remarkably tight risk certificates, bearing promise towards self-certified learning (\emph{i.e.} use all the data to learn a predictor and certify its quality).
In this work, we empirically investigate the role of the prior.
We experiment on 6 datasets with different strategies and amounts of data to learn data-dependent PAC-Bayes priors, and we compare them in terms of their effect on test performance of the learnt predictors and tightness of their risk certificate.
We ask what is the optimal amount of data which should be allocated for building the prior and show that the optimum may be dataset dependent. 
We demonstrate that using a small percentage of the prior-building data for validation of the prior leads to promising results.
We include a comparison of underparameterised and overparameterised models, along with an empirical study of different training objectives and regularisation strategies to learn the prior distribution. 
\end{abstract}


\section{Introduction}

The majority of deep learning algorithms output weights of neural networks: in recent years, a growing body of works has investigated algorithms which rather output \emph{probability distributions} over the connection weights of a network, with a number of advantages.
This is evidenced \emph{e.g.} by works 
inspired by Bayesian learning (see e.g. \cite{blundell2015weight,OsawaSKJETY2019,Wilson20} among many others) or by the frequentist PAC-Bayes bounds (see e.g. \cite{dziugaite2017computing,letarte2019dichotomize,biggs2020,perez2020}).
In both cases, a probability distribution over neural network weights defines what can be called a Probabilistic Neural Network (PNN). 


Recently, PNNs learnt by optimising PAC-Bayes bounds have shown promising results on performance guarantees, by delivering tight risk certificates (generalisation bounds) for predictive models that are competitive compared to standard empirical risk minimisation (see \cite{perez2020}).
Importantly, the PNN paradigm, coupled with PAC-Bayes bounds, is an example of self-certified learning,
which proposes to use all the available data for learning a predictor and providing a reasonably tight numerical risk bound value that certifies the predictor's performance at the population level.
In this case, the risk certificates can be evaluated on a subset of the data used for training and thus do not require a held-out test set, allowing efficient use of the available data.
%
These principled learning and certification strategies based on PAC-Bayes bounds deserve further study to unfold their practical properties and limitations. 



PAC-Bayes bounds (pioneered by \cite{Shawe-TaylorWilliamson1997,McAllester1998,McAllester1999}) are typically composed of two key quantities: i) a term that measures the empirical performance of a so-called `posterior' distribution, and ii) a term involving the divergence of the posterior to a `prior' distribution, which in most bounds is the Kullback-Leibler (KL) divergence (we refer to \cite{guedj2019primer} for a comprehensive presentation). 
Thus, when using a PAC-Bayes bound as an optimisation objective, these two terms must interact to balance fitness to data with fitness to the chosen prior. 
A classical assumption underlying PAC-Bayes priors is that they must be independent from the data on which the empirical term of the PAC-Bayes bound is evaluated. 
This assumption is well-known in the PAC-Bayes literature (as discussed by \cite{catoni2004,catoni2007}).
Interestingly, however, the chosen prior greatly impacts the  bound via the KL term, which often amounts to the dominating contribution to bound values (as pointed out by \cite{dziugaite2021role,perez2020}).
This prominent role of the KL term has implications both for optimisation and for risk certification based on PAC-Bayes bounds: (i) the KL term effectively constrains the posterior such that it cannot move too far from the prior; (ii) the large values of the KL term when using data-independent priors or uninformed priors suggests that these priors may not be able to give tight risk certificates, therefore calling for data-dependent priors.

These considerations give clues on the importance of the prior distribution in PAC-Bayes bounds:
An arbitrarily chosen prior may mislead the optimisation, since the posterior is constrained to the prior by the KL term, while a prior representing a good solution for the problem at hand would give a better starting point.
Accordingly, some works on PAC-Bayes bounds for neural networks have considered ways to connect PAC-Bayes priors to the data. 
In particular, the recent works in \cite{dziugaite2021role} and \cite{perez2020} explored PAC-Bayes priors learnt on a subset of the data, which does not overlap with the data used for computing the empirical term in the PAC-Bayes bound.
This way, these data-dependent PAC-Bayes priors are in line with the classical assumption underlying PAC-Bayes priors, while at the same time the priors have a more sensible connection to the data when compared to arbitrarily chosen ones.


In this work, we investigate strategies for learning the PAC-Bayes prior, and the role of the prior in obtaining  PNNs with good performance by optimising PAC-Bayes bounds.
\newpage




\paragraph{Our Contributions.}

\begin{itemize}
\item We demonstrate consistency of the tightness of the risk certificates across $6$ classification datasets. 
\item We contribute supporting evidence on the crucial role of learning the PAC-Bayes prior for achieving efficient neural network classifiers and tight risk certificates. 
\item We show that there is potential for tight risk certificates even for highly over-parameterised architectures.

\end{itemize}

\section{Related Literature}

\noindent\textbf{Bayesian Neural Networks (BNNs).} BNNs are special cases of PNNs, 
in the sense that BNN methods also output distributions over network weights.
Works from this front have shown very promising results, e.g. that BNNs provide an intuitive approach to uncertainty quantification and principled implementations of model pruning/distillation \cite{blundell2015weight}. 
Furthermore, training objectives inspired by the evidence lower bound (ELBO) and commonly used in BNNs have been seen to act as an implicit regulariser.
The resemblance between the ELBO and PAC-Bayes bounds, which has been pointed out in the literature (\cite{AlquierRC2016,AchilleSoatto2018},
among others), suggests that some of these properties may be shared by methods inspired by PAC-Bayes bounds.



\noindent\textbf{Bayes vs. PAC-Bayes.}
The `prior' and `posterior' distributions that appear in PAC-Bayes bounds should not be confused with their Bayesian counterparts. 
In PAC-Bayes bounds, what is called `prior' is a reference distribution, and what is called `posterior' is an unrestricted distribution, in the sense that there is no likelihood factor connecting them  (cf. \cite{guedj2019primer,rivasplata2020beyond}).

\noindent\textbf{Data-dependent PAC-Bayes priors.}
In this line of thought, a precursor work is that in 
\cite{dziugaite2018data}. 
The protocol for partitioning the data goes back to the works in \cite{ambroladze2007tighter,lever-etal2010,JMLR:v13:parrado12a,lever-etal2013}.


\section{Elements of Statistical Learning}

Supervised classification algorithms receive training data $S = ((X_1,Y_1),\ldots,(X_n,Y_n))$ consisting of pairs that encode inputs $X_i \in \cX \subseteq \R^d$ and their labels $Y_i \in \cY$. 
Classifiers $h_w : \cX \to \cY$ are mappings from input space $\cX$ to label space $\cY$, and we assume they are parametrised by `weight vectors' $w \in \cW \subseteq \R^p$.
The quality of $h_w$ is given by its risk $L(w)$, which by definition is the expected classification error on a randomly chosen pair $(X,Y)$. 
However, $L(w)$ is an inaccessible measure of quality, since the distribution that generates the data is unknown.
An accessible measure of quality is given by the empirical risk functional
$\hL_S(w)
    = n^{-1}\sum_{i=1}^{n} \ell(h_w(X_i),Y_i)$,
defined in terms of a loss function $\ell : \R \times \cY \rightarrow [0,\infty)$. 
Indeed, the empirical risk minimisation (ERM) paradigm aims to find $w \in \cW$ that minimises this functional for some choice of loss function (zero-one loss or a surrogate loss).


The outcome of training a PNN is a distribution $Q$ over weight space and this distribution depends on the sample $S$.
Then, given a fresh input $X$, the randomised classifier predicts its label by drawing a weight vector $W$ at random from $Q$ and applying the predictor $h_W$ to $X$.
For the sake of simplicity, we identify the randomised predictor with the distribution $Q$ that defines it.
The quality of this randomised predictor is measured by the expected loss notions under the random draws of weights. 
Thus, the loss of $Q$ is given by 
$Q[L] = \int_{\cW} L(w) Q(dw)$;
and the empirical loss of $Q$ is given by
$Q[\hL_S] = \int_{\cW} \hL_S(w) Q(dw)$. 

\begin{figure*}[ht]
\begin{center}
    \centerline{\includegraphics[width=0.9\textwidth]{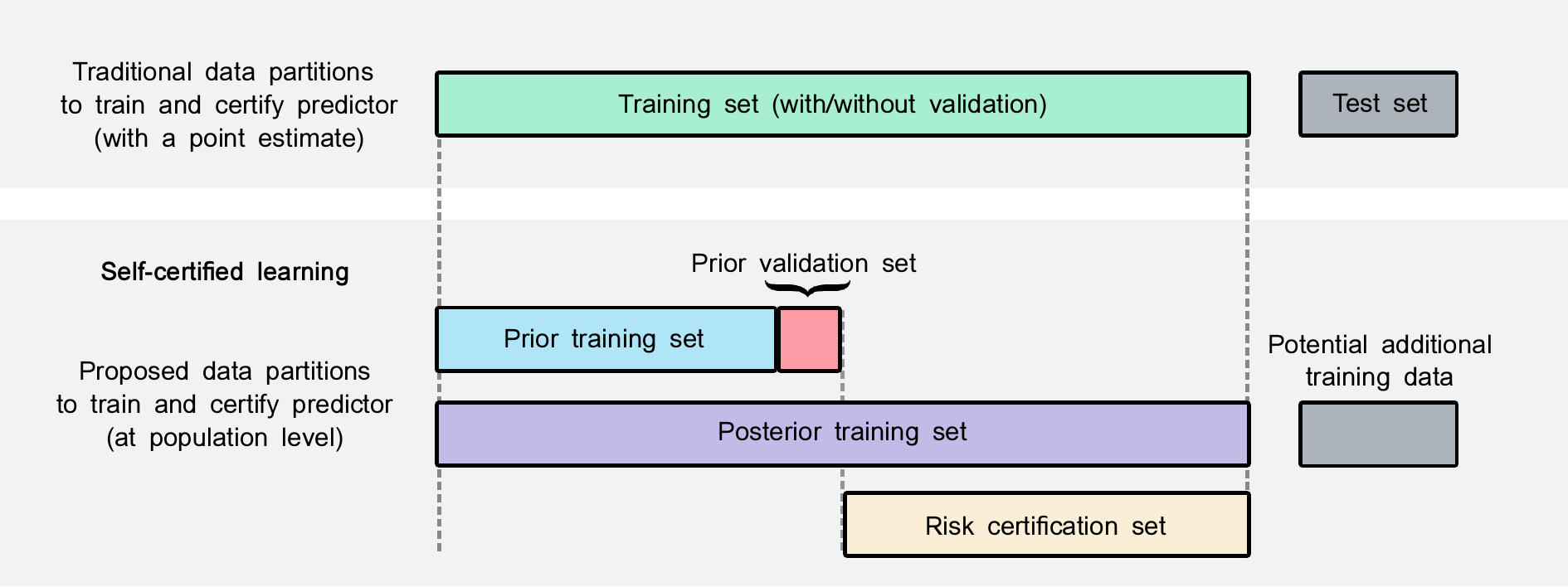}}
    \caption{Comparison of data partitions in traditional machine learning (where the model is certified with a point estimate) vs self-certified learning through PAC-Bayes bounds. Note
that all $n$ training data are used by the self-certified learning algorithm.} 
    \label{fig:partitions}
\end{center}
\end{figure*}

The PAC-Bayes-quadratic bound in \cite{perez2020} says that for any $\delta \in (0,1)$,
with probability of at least $1-\delta$ over size-$n$ i.i.d. random samples $S$,
simultaneously for all distributions $Q$ over $\cW$ we have:
\begin{equation}
\label{eq:quad-bound}
     Q[L] 
    \leq  \left(
    \sqrt{ 
    Q[\hL_S] + \frac{\KL(Q \Vert Q^0) + \log(\frac{2\sqrt{n}}{\delta})}{2n} 
    } \right. \nonumber
    +  \left.
    \sqrt{ 
    \frac{\KL(Q \Vert Q^0) + \log(\frac{2\sqrt{n}}{\delta})}{2n} 
    } \right)^2. 
\end{equation}
In this case the prior $Q^0$ must be chosen without any dependence on the data $S$ on which the empirical term $\hL_S$ is evaluated. In this work, we use a partitioning scheme for the training data $S = S_{\mathrm{pri}} \cup S_{\mathrm{cert}}$ such that the prior is trained on $S_{\mathrm{pri}}$, the posterior is trained on the whole set $S$ and the risk certificate is evaluated on $S_{\mathrm{cert}}$. See \figurename{~\ref{fig:partitions}}.

\section{Learning and Certification Strategy}

In a nutshell, the learning and certification strategy has three components:
(1) choose/learn a prior;
(2) learn a posterior; and
(3) evaluate the risk certificate for the posterior.

\subsection{Data-dependent PAC-Bayes priors}

We experiment with Gaussian priors $Q^0$ with a diagonal covariance matrix centered at i) random weights (uninformed data-free priors) and ii) learnt weights (data-dependent priors) on a subset of the dataset which is independent of the subset used to compute the risk certificate (see \figurename{~\ref{fig:partitions}}). 
In all cases, the posterior is initialised to the prior. Similar approaches have been considered before in the PAC-Bayesian literature
(we refer to \cite{
JMLR:v13:parrado12a,
perez2020,
dziugaite2021role
}.

\subsubsection{Training objectives to learn the prior}

We evaluate several training strategies to learn the PAC-Bayes prior: 
\begin{enumerate}
    \item Deterministic training objectives: 
    We compare learning the prior mean by non-regularised ERM (referred to as $f_{\texttt{erm}}$) to ERM with dropout ($f_{\texttt{erm}}$ + dropout) and Mixup \cite{mixup} ($f_{\texttt{mixup}}$), which has been shown to act as an implicit regulariser \cite{mixupreg}. Note that all of these training objectives learn a deterministic neural network with a fixed setting of weights. The prior distribution is then centered at the learnt weight vector, and the scale of the distribution is set as a hyper-parameter.
    \item Probabilistic training objectives: 
    We compare learning the prior distribution via the Bayes by backprop ($f_{\texttt{bbb}}$) \cite{blundell2015weight} objective inspired by the evidence lower bound, to learning it via the  objective inspired by the PAC-Bayes-quadratic bound ($f_{\texttt{quad}}$) \cite{perez2020}. 
    Since these objectives are probabilistic, they learn the full prior distribution, and not only the mean, as it is the case with the deterministic objectives. 
    For both of these objectives, 
    we first need to set a pre-prior, for which we use uninformed priors centered at zero or at a random weights, setting the scale as a hyper-parameter. We then use this uninformed prior in the KL computation of the training objective, in order to learn the data-dependent prior, which is later used to train the posterior. 
\end{enumerate}

\subsubsection{Validation of prior and posterior}\label{sec:validation}

The work in \cite{perez2020} highlighted the difficulty of avoiding overfitting of the prior learnt by ERM and suggested the use of dropout. In this paper, we propose to use the following strategy to monitor the performance of prior and posterior and choose the best prior/posterior accordingly:
\begin{enumerate}
    \item Prior validation: We propose to select a small percentage of the prior-building data to use as a validation set, with the objective of monitoring the validation loss while learning the prior predictor. Then, at the end of the prior-building stage we choose the one that performs the best in the validation set.  
    \item Posterior validation: In this case we compute the risk certificate at different stages during posterior learning and choose the posterior with the best risk certificate, thus not needing any additional data partition. 
\end{enumerate}
A summary of the partitions used at the different stages of learning and certification is shown in \figurename{~\ref{fig:partitions}}. Note that the prior validation set is still being used to train the posterior, hence not discarding any data from the learning process.

\subsection{Posterior Optimisation \& Certification}

We now present the essential idea of training PNNs by minimizing a PAC-Bayes upper bound on the risk. 
We use a recently proposed PAC-Bayes inspired training objective \cite{perez2020}, derived from Eq.~\eqref{eq:quad-bound} 
in the context of neural network classifiers:
\begin{equation}
\label{eq:obj-quad}    
     f_{\mathrm{quad}}(Q) 
     = 
    \left(
    \sqrt{ 
    Q[\hL^{\ce}_S] + \frac{\KL(Q \Vert Q^0) + \log(\frac{2\sqrt{n}}{\delta})}{2n} 
    } \right. \nonumber
+ \left.
    \sqrt{ 
    \frac{\KL(Q \Vert Q^0) + \log(\frac{2\sqrt{n}}{\delta})}{2n} 
    } \right)^2. 
\end{equation}
This objective is implemented using the cross-entropy loss, which is
the standard surrogate loss commonly used on these problems. 
Since the PAC-Bayes bounds of Eq.~\eqref{eq:quad-bound} require the loss within [0,1], we construct a `bounded cross-entropy' loss by lower-bounding the network probabilities by a value $p_\mathrm{min}>0$ (cf. \cite{dziugaite2018data, perez2020}). 
The empirical risk term is $\hL^{\ce}_S(w)$ is calculated with the `bounded' version of the cross-entropy loss.

Optimization of the objective in Eq.  \eqref{eq:obj-quad} entails minimizing over $Q$.
By choosing $Q$ in a parametric family of distributions, we can use the pathwise gradient estimator 
\cite{price1958useful,jankowiak2018pathwise}
as done by \cite{blundell2015weight}.
The details of the reparametrisation strategy are outlined in \cite{perez2020}.
Following \cite{blundell2015weight}, the reparametrization we use is $W = \mu + \sigma \odot V$ with Gaussian distributions for each coordinate of $V$. The optimization uses $\sigma = \log(1+\exp(\rho))$, thus gradient updates are with respect to $\mu$ and $\rho$.

\paragraph{Evaluation of the Risk Certificates.}  After optimising the posterior distribution over network weights through the previously presented training objective, we compute a risk certificate on the error of the stochastic predictor.
To do so, we follow the procedure outlined in \cite{perez2020}, which was used before by \cite{dziugaite2017computing} and goes back to the work of \cite{LangfordCaruana2001}. This certification procedure uses the PAC-Bayes-kl bound.
In particular, the procedure is based on numerical inversion of the binary KL divergence, as done by \cite{dziugaite2017computing,perez2020}. 



\section{Experiments}

We conduct the following five experiments on the effect of PAC-Bayes priors on PNN predictors: 
\begin{enumerate}
    \item[E1] We study the relationship between the learnt PAC-Bayes prior predictive performance and the posterior risk certificate in a large set of experiments on MNIST. 
    \item[E2] We test if the validation of prior and posterior leads to better performance, even when the amount of data used for prior learning is effectively reduced. 
    \item[E3] We experiment with different trade-offs of the amount of data to learn the prior and certify the predictor, extending the results in \cite{perez2020} to 5 additional datasets, which demonstrates the role of data-dependent priors on the tightness of the risk certificates. 
        \item[E4] We study the role of the number of parameters in the architecture in the risk certificate and KL term. 
    \item[E5] Finally, we compare several training objectives and regularisation strategies for learning the prior. 
\end{enumerate}


Table \ref{tab:datasets} shows the datasets used (all except for MNIST available at OpenML.org), selected so as to represent a wide range of characteristics (small vs large, low vs high dimensional and binary vs multiclass). For all datasets except MNIST we select 20\% of the data as test set (stratified with class label). For MNIST, we use the standard data partitions. 

\begin{table}[ht!]
\centering
\small
\setlength{\tabcolsep}{3pt}
 \begin{tabular}{| c | cc c|} 
 \hline
Dataset	&	$n$	&	$\#f$ 	&	$\#c$	\\ \hline
Spambase	&	4601	&	58	&	2	\\
Bioresponse	&	3751	&	1777	&	2	\\
Har	&	10299	&	562	&	6	\\
Electricity	&	45312	&	9	&	2	\\
Mammography	&	11183	&	7	&	2	\\
MNIST	&	70000	&	784	&	10	\\
 \hline
\end{tabular} 
\caption{Datasets used: $n$ is the total number of data points, $\#f$ the number of features and $\#c$ the number of classes. 
}
\label{tab:datasets}
\end{table}

\begin{table*}[htb]
\centering
\scriptsize
\setlength{\tabcolsep}{2.5pt}
 \begin{tabular}{| c c | c c cc| cc |} 
 \hline
	&		&	\multicolumn{4}{c}{Posterior}				&		\multicolumn{2}{|c|}{Prior}			\\ \hline
Dataset	&	Val.	&	R. Cert.	&	S. 01 err.	&	KL/$n_{\texttt{cert}}$ & Post. mean 01 err. &	Prior mean 01 err.	&	S. 01 err. 	\\ \hline
\multirow{2}{*}{Spambase ($n_{\texttt{cert}}=1840$)}	&	\xmark	&$	0.140	$&$	0.082	$&$	5e^{-5}	$&$	0.077	$&$	0.077	$&$	0.083	$\\
	&	\cmark	&$	0.127	$&$	0.065	$&$	3e^{-4}	$&$	0.059	$&$	0.056	$&$	0.067	$\\ \hline
\multirow{2}{*}{Bioresponse ($n_{\texttt{cert}}=1500$)}	&	\xmark	&$	0.318	$&$	0.267	$&$	9e^{-4}	$&$	0.262	$&$	0.261	$&$	0.268	$\\
	&	\cmark		&$	0.291	$&$	0.257	$&$	5e^{-5}	$&$	0.250	$&$	0.248	$&$	0.258	$\\ \hline
\multirow{2}{*}{Har ($n_{\texttt{cert}}=4119$)}	& \xmark &$	0.035	$&$	0.021	$&$	3e^{-4}	$&$	0.017	$&$	0.024	$&$	0.026	$\\
	&	\cmark		&$	0.037	$&$	0.024	$&$	8e^{-6}	$&$	0.021	$&$	0.020	$&$	0.025	$\\ \hline
\multirow{2}{*}{Electricity ($n_{\texttt{cert}}=18124$)}	&	\xmark		&$	0.223	$&$	0.214	$&$	6e^{-5}	$&$	0.203	$&$	0.221	$&$	0.227	$\\
	&	\cmark		&$	0.221	$&$	0.212	$&$	3e^{-5}	$&$	0.203	$&$	0.205	$&$	0.213	$	\\ \hline
\multirow{2}{*}{Mammography ($n_{\texttt{cert}}=4473$)}	&	\xmark	&$	0.022	$&$	0.015	$&$	3e^{-6}	$&$	0.015	$&$	0.015	$&$	0.016	$\\
	&	\cmark	&$	0.023	$&$	0.017	$&$	3e^{-7}	$&$	0.017	$&$	0.017	$&$	0.017	$\\
 \hline
 \multirow{2}{*}{MNIST  ($n_{\texttt{cert}}=30000$)}	&	\xmark &$	0.034	$&$	0.026	$&$	1e^{-5}	$&$	0.024	$&$	0.025	$&$	0.027	$\\
	&	\cmark	&$0.030		$&$	0.027	$&$	2e^{-5}	$&$	0.027	$&$	0.028	$&$	0.031	$\\ \hline
\end{tabular} 
\caption{Results with/without the validation set to choose the prior/posterior in an experiment using data-dependent priors. The stochastic 01 test set errors (S. 01 err.) for both prior and posterior are averaged over 10 samples. We do not include the standard deviation because they are low (around 0.001 in 6 cases and lower than 0.001 in the rest). The table also includes the 01 error of the posterior and prior mean. $n_\texttt{cert}$ indicates the number of data points used for the computation of the risk certificate.
}
\label{tab:validation}
\end{table*}

\begin{table*}[htb]
\centering
\scriptsize
\setlength{\tabcolsep}{2.5pt}
\begin{tabular}{| c | cc|c | cc|c | cc|c|} 
 \hline
Dataset	&	\multicolumn{3}{|c|}{Spambase}					&		\multicolumn{3}{|c|}{Bioresponse}					&		\multicolumn{3}{|c|}{Har}			
\\ \hline
		&	\multicolumn{2}{|c|}{Posterior} & \multicolumn{1}{|c|}{Prior}					&		\multicolumn{2}{|c|}{Posterior} & \multicolumn{1}{|c|}{Prior}				&	\multicolumn{2}{|c|}{Posterior} & \multicolumn{1}{|c|}{Prior}						
\\ \hline
Perc. of data for prior	&	R. Cert.	&	 S. 01 err. 	& S. 01 err. &	R. Cert. 	&	 S. 01 err. 	 & S. 01 err. 	&	R. Cert.	& S. 01 err.	 	 & S. 01 err.
\\ \hline
Data-free (0\%)	& $0.485 $ & $ 0.406 $ & $ 0.545 $ & $ 0.531 $ & $ 0.494 $ & $ 0.505 $ & $ 0.789 $ & $ 0.758 $ & $ 0.759 $ \\
Data-depend. (10\%) & $0.213 $ & $ 0.153 $ & $ 0.182 $ & $ 0.344 $ & $ 0.315 $ & $ 0.315 $ & $ 0.072 $ & $ 0.063 $ & $ 0.077 $ \\
Data-depend. (25\%)	&$0.197 $ & $ 0.133 $ & $ 0.100 $ & $ 0.314 $ & $ 0.270 $ & $ 0.266 $ & $ 0.042 $ & $ 0.036 $ & $ 0.038 $ \\
Data-depend. (50\%)	&$\mathbf{0.127} $ & $ 0.065 $ & $ 0.067 $ & $ \mathbf{0.291} $ & $ 0.258 $ & $ 0.257 $ & $ 0.037 $ & $ 0.024 $ & $ 0.025 $ \\
Data-depend. (75\%)	& $0.144 $ & $ 0.056 $ & $ 0.062 $ & $ 0.305 $ & $ 0.258 $ & $ 0.258 $ & $\mathbf{0.035} $ & $ 0.023 $ & $ 0.024 $ \\
Data-depend. (90\%)	& $0.182 $ & $ 0.062 $ & $ 0.065 $ & $ 0.345 $ & $ 0.255 $ & $ 0.255 $ & $ 0.037 $ & $ 0.018 $ & $ 0.018 $ \\
\hline
 ERM baseline (01 err.) & \multicolumn{3}{|c|}{$0.045$}  &			\multicolumn{3}{|c|}{$0.229$}		&	\multicolumn{3}{|c|}{$0.015$}	
\\ \hline
Dataset &		\multicolumn{3}{|c|}{Electricity}					&		\multicolumn{3}{|c|}{Mammography}	&		\multicolumn{3}{|c|}{MNIST}							
\\ \hline
&		\multicolumn{2}{|c|}{Posterior} & \multicolumn{1}{|c|}{Prior}							&		\multicolumn{2}{|c|}{Posterior} & \multicolumn{1}{|c|}{Prior}			&		\multicolumn{2}{|c|}{Posterior} & \multicolumn{1}{|c|}{Prior}							
\\ \hline
Perc. of data for prior	&	R. Cert. 	&	 S. 01 err. 	& S. 01 err. &	R. Cert. 	& S. 01 err.	 		& S. 01 err. & R. Cert. 	& S. 01 err.		& S. 01 err.	
\\ \hline
Data-free (0\%) & $0.441 $ & $ 0.427 $ & $ 0.427 $ & $ 0.038 $ & $ 0.023 $ & $ 0.214 $ & $ 0.321 $ & $ 0.220 $ & $ 0.891 $\\
Data-depend. (10\%)& $0.244 $ & $ 0.234 $ & $ 0.235 $ & $ 0.024 $ & $ 0.020 $ & $ 0.019 $ & $ 0.082 $ & $ 0.075 $ & $ 0.076 $ \\
Data-depend. (25\%)	& $0.234 $ & $ 0.224 $ & $ 0.232 $ & $ 0.024 $ & $ 0.018 $ & $ 0.018 $ & $ 0.047 $ & $ 0.046 $ & $ 0.047 $ \\
Data-depend. (50\%)	& $0.233 $ & $ 0.212 $ & $ 0.227 $ & $ \mathbf{0.023} $ & $ 0.017 $ & $ 0.017 $ & $ 0.030 $ & $ 0.027 $ & $ 0.031$ \\
Data-depend. (75\%)	& $\mathbf{0.028} $ & $ 0.023 $ & $ 0.208 $ & $ 0.027 $ & $ 0.018 $ & $ 0.018 $ & $ 0.028 $ & $ 0.023 $ & $ 0.029 $\\
Data-depend. (90\%)	& $0.223 $ & $ 0.201 $ & $ 0.202 $ & $ 0.031 $ & $ 0.017 $ & $ 0.017 $ & $ \mathbf{0.026} $ & $ 0.022 $ & $ 0.026 $\\
 \hline
 ERM baseline (01 err.)  &		\multicolumn{3}{|c|}{$0.192$}	 &		\multicolumn{3}{|c|}{$0.017$}  &		\multicolumn{3}{|c|}{$0.022$}
\\ \hline
\end{tabular} 
\caption{Results when using different percentages of the dataset for prior building and evaluation of the risk certificate. Best risk certificates for each dataset are highlighted in bold face.}
\label{tab:percprior}
\end{table*}

\begin{table*}[htb]
\centering
\scriptsize
\setlength{\tabcolsep}{2pt}
 \begin{tabular}{| c | cc|cc | cc|cc | cc|} 
 \hline
\multicolumn{1}{|c|}{Dataset}	&	\multicolumn{2}{|c|}{Spambase}					&		\multicolumn{2}{|c|}{Bioresponse}					&		\multicolumn{2}{|c|}{Har}		&		\multicolumn{2}{|c|}{Electricity}					&		\multicolumn{2}{|c|}{Mammography}			
\\ \hline
Prior objective	&		R. Cert. 	&	 S. 01 err. 	& 	R. Cert. 	&	 S. 01 err. 	& 	R. Cert. 	&	 S. 01 err. 	& 	R. Cert. 	&	 S. 01 err. 	& 	R. Cert. 	&	 S. 01 err. 	
\\ \hline									
$f_{\texttt{erm}}$	&$	0.132	$&$	0.088	$&$	0.291	$&$	0.270	$&$	0.036	$&$	0.024	$&$	0.222	$&$	0.215	$&$	0.022	$&$	0.017	$
\\	
$f_{\texttt{erm}}$ + dropout	&$	0.126	$&$	0.063	$&$	0.290	$&$	0.255	$&$	0.033	$&$	0.021	$&$	0.219	$&$	0.205	$&$	0.022	$&$	0.017	$
\\
$f_{\texttt{mixup}}$	&$	0.166	$&$	0.105	$&$	0.299	$&$	0.272	$&$	0.038	$&$	0.019	$&$	0.241	$&$	0.231	$&$	0.028	$&$	0.021	$
\\ \hline
$f_{\texttt{bbb}}$ (random prior)	&$	0.123	$&$	0.068	$&$	0.290	$&$	0.249	$&$	0.031	$&$	0.020	$&$	0.222	$&$	0.209	$&$	0.022	$&$	0.017	$
\\
$f_{\texttt{bbb}}$ (zero prior)	&$	0.120	$&$	0.060	$&$	0.288	$&$	0.254	$&$	0.031	$&$	0.020	$&$	0.214	$&$	0.204	$&$	0.022	$&$	0.017	$
\\
$f_{\texttt{quad}}$ (random prior)	&$	0.186	$&$	0.129	$&$	0.309	$&$	0.272	$&$	0.045	$&$	0.033	$&$	0.247	$&$	0.237	$&$	0.027	$&$	0.018	$ 
\\
$f_{\texttt{quad}}$ (zero prior)	&$	0.243	$&$	0.168	$&$	0.309	$&$	0.272	$&$	0.045	$&$	0.033	$&$	0.247	$&$	0.237	$&$	0.027	$&$	0.018	$ 
\\ \hline
\end{tabular} 
\caption{Results with different training objectives to learn the prior. $f_{\texttt{erm}}$ and $f_{\texttt{mixup}}$ based objectives are deterministic and only learn the mean of the prior distribution. $f_{\texttt{bbb}}$ and $f_{\texttt{quad}}$ are probabilistic and learn the full prior distribution. 
}
\label{tab:objects}
\end{table*}

\subsection{Experimental setup}

 In all experiments the models are compared under the same conditions, i.e. weight initialisation and optimiser (vanilla SGD with momentum). The mean parameters $\mu_0$ of the prior are initialised randomly from a truncated centered Gaussian distribution with standard deviation set to $1/\sqrt{n_\mathrm{in}}$, where $n_\mathrm{in}$ is the dimension of the inputs to a particular layer, truncating at $\pm 2$ standard deviations. 
All risk certificates are computed using the the PAC-Bayes-kl inequality, as explained in Section 6 of \cite{perez2020}, with $\delta=0.025$ and $\delta'=0.01$  and $m=150 000$ Monte Carlo model samples.
We also report the average 01 error of the stochastic predictor, where we randomly sample fresh model weights for each test example $10$ times and compute the average 01 error. 
Input data was standardised for all datasets. 

For all experiments we performed a grid search over all hyper-parameters and selected the run with the best risk certificate. We did a grid sweep over the prior distribution scale hyper-parameter (i.e. standard deviation $\sigma_0$) with values in $[0.1, 0.05, 0.04, 0.03, 0.02, 0.01]$.  For SGD with momentum we performed a grid sweep over learning rate in $[1\mathrm{e}^{-3}, 5\mathrm{e}^{-3}, 1\mathrm{e}^{-2}]$ and momentum in $[0.95, 0.99]$. We also performed a grid sweep over the learning rate and momentum used for learning the prior (testing the same values as before). The dropout rate used for learning the prior was selected from $[0.01, 0.05, 0.1, 0.2]$. 
 The grid sweep is done for experiment E2. In the subsequent experiments, we use the same best performing hyper-parameters from E2.
We experiment with fully connected neural networks (FCN) with 2/3 layers (excluding the `input layer') and 100 units per hidden layer (unless specified otherwise).  
ReLU activations are used in each hidden layer.
For learning the prior we ran the training for 500 epochs (except for MNIST for which we ran 100). Posterior training was run for 100 epochs. We use a training batch size of $250$. PyTorch code will be released anonymously at the repository associated to this project\footnote{\url{https://anonymous.4open.science/r/pacbayespriors-F355/}}.

\subsection{E1: The role of the PAC-Bayes prior in PNNs}

\begin{figure*}[ht!]
\begin{center}
    \centerline{\includegraphics[width=\textwidth]{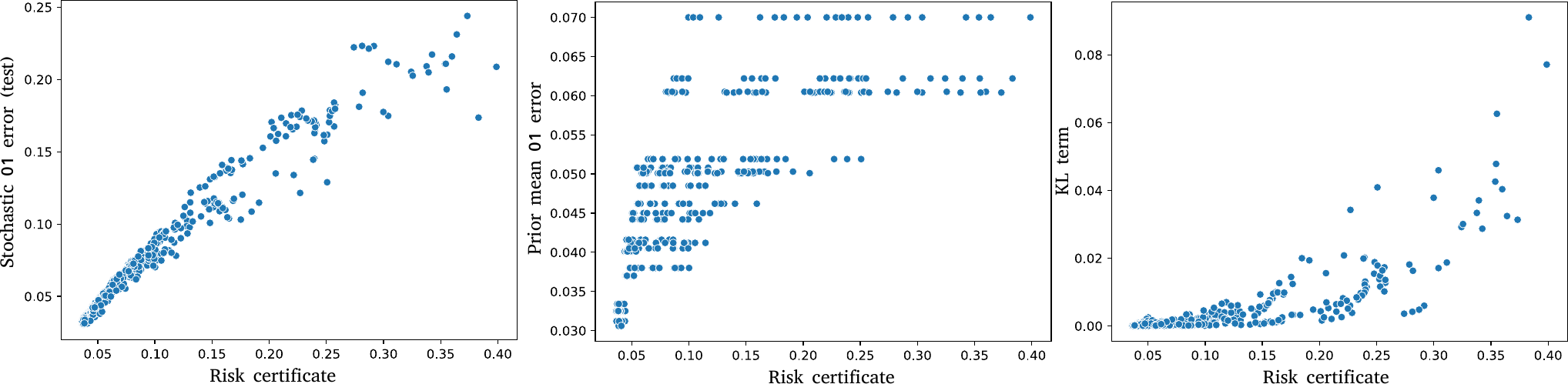}}
    \caption{Comparison of several metrics vs the risk certificate for more than 600 runs with different hyper-parameters on MNIST as done in \cite{perez2020}.
We use a reduced subset of MNIST for these experiments (10\% of training data).  
    }
    \label{fig:roleprior}
\end{center}
\end{figure*}

\begin{table}[ht]
\centering
\scriptsize
\setlength{\tabcolsep}{2pt}
 \begin{tabular}{| c | ccc | ccc| } 
 \hline
 	&	\multicolumn{3}{|c|}{FCN Architecture} & \multicolumn{3}{|c|}{Posterior} \\ \hline	
Dataset  & \#Params & Neurons & H. Lay. & R. Cert. & S. 01 err. & KL/$n_\texttt{cert}$ \\ \hline
\multirow{4}{*}{Spamb.}  & $1404$ & $10$ & $2$ & $0.130$ & $0.088$ & $1e^{-4}$  \\
  & $32004$ & $100$ & $2$ & $0.132$ & $0.088$ & $9e^{-5}$ \\
  & $792004$ & $600$ & $2$ & $0.120$ & $0.060$ & $4e^{-4}$\\ 
  & $1512004$ & $600$ & $3$ & $0.120$ & $0.060$ & $6e^{-4}$ \\ \hline
\multirow{5}{*}{Bioresp.}  & $7128$ & $3$ & $2$ & $0.521$ & $0.483$ & $3e^{-8}$  \\
  & $35784$ & $10$ & $2$ & $0.287$ & $0.254$ & $9e^{-6}$ \\
  & $375804$ & $100$ & $2$ & $0.291$ & $0.270$ & $4e^{-5}$  \\
  & $395804$ & $100$ & $3$ & $0.290$ & $0.248$ & $2e^{-5}$  \\
  & $2854804$ & $600$ & $2$ & $0.283$ & $0.257$ & $3e^{-5}$ \\  \hline
\multirow{5}{*}{Har}  & $4588$ & $4$ & $2$ & $0.405$ & $0.376$ & $7e^{-6}$  \\
  & $11572$ & $10$ & $2$ & $0.040$ & $0.024$ & $3e^{-5}$ \\
  & $133612$ & $100$ & $2$ & $0.036$ & $0.024$ & $9e^{-6}$  \\
  & $153612$ & $100$ & $3$ & $0.037$ & $0.027$ & $5e^{-5}$ \\
  & $1401612$ & $600$ & $2$ & $0.038$ & $0.023$ & $3e^{-5}$ \\ \hline
\multirow{4}{*}{Elect.}  & $424$ & $10$ & $2$ & $0.242$ & $0.230$ & $7e^{-6}$ \\
  & $22204$ & $100$ & $2$ & $0.222$ & $0.215$ & $2e^{-5}$ \\
  & $733204$ & $600$ & $2$ & $0.220$ & $0.209$ & $1e^{-5}$ \\
  & $1453204$ & $600$ & $3$ & $0.211$ & $0.203$ & $6e^{-5}$  \\ \hline
\multirow{4}{*}{Mammo.}  & $384$ & $10$ & $2$ & $0.026$ & $0.018$ & $3e^{-7}$ \\
  & $21804$ & $100$ & $2$ & $0.022$ & $0.017$ & $4e^{-7}$ \\
  & $730804$ & $600$ & $2$ & $0.025$ & $0.018$ & $6e^{-5}$ \\
  & $1450804$ & $600$ & $3$ & $0.027$ & $0.018$ & $9e^{-5}$ \\\hline
\multirow{4}{*}{MNIST}  & $16100$ & $10$ & $2$ & $0.071$ & $0.068$ & $2e^{-4}$ \\
  & $178820$ & $100$ & $2$ & $0.030$ & $0.027$ & $2e^{-4}$ \\
  & $1672820$ & $600$ & $2$ & $0.028$ & $0.023$ & $9e^{-5}$ \\
  & $2392820$ & $600$ & $3$ & $0.029$ & $0.024$ & $2e^{-4}$ \\\hline
\end{tabular} 
\caption{Results with architectures with different numbers of parameters (both under and over parametrised with respect to the dataset size \cite{Belkin15849}). 
}
\label{tab:archs}
\end{table}

The work in 
\cite{perez2020} showed that PAC-Bayes bounds can be used not only as training objectives but also for model selection. Specifically, the authors compared the risk certificates obtained  to the test stochastic 01 error in a large-scale hyper-parameter grid search experiment involving more than 600 runs on MNIST,
showing a strong relationship which motivated using the risk certificates for model selection.

We begin by replicating this experiment to study the role of the prior on the risk certificate.
%
To do this, we perform the same grid search with the architecture reported in \cite{perez2020} over 6 hyper-parameters: prior scale, dropout rate, and the learning rate and momentum both for learning the prior and the posterior. These experiments use a reduced MNIST training set (i.e. 10\%), while the test set is  maintained. 
\figurename{~\ref{fig:roleprior}} shows the relationship between three metrics of interest and the risk certificate.  We start by comparing the test 01 loss of the stochastic predictor to the risk certificate as done in \cite{perez2020} (right plot). We observe a similarly strong relationship. 
The  middle plot shows the relationship between 
the deterministic test 01 loss of the prior mean learnt by ERM and the risk certificate.
Interestingly, this plot shows that when the prior is competitive, then the risk certificate is as well. However, good posteriors and risk certificates can be achieved even if the prior is not very competitive. Studying the hyper-parameters in this experiment, we observe that both the standard deviation used in the prior and the use of dropout when learning the prior have a big impact on the posterior risk certificate, independently of the prior's test set 01 error. Finally, 
the right part of the plot shows the relationship between the KL term in the bound and the risk certificate. This plot demonstrates the significant influence of the KL term (and thus the prior) in the final risk certificate.

\subsection{E2: Validation of prior and posterior}

We now test the use of the validation strategy to choose the best fitted prior and posterior. 
To do so, we use 50\% of the training data to build the prior, and 5\% of those 50\% for validation.
The results of this experiment can be seen in \tablename{ \ref{tab:validation}}, where the column named \emph{Validation} indicates whether we use the above mentioned strategy to select the best prior and posterior. 
We note that even when discarding a small part of the prior set, this strategy improves the risk certificate in most cases (4 out of 6 datasets). 
 This seems to lead to much smaller values of the KL term.
However, the posterior stochastic 01 error does not improve in all cases (worsens in 3 out of the 6 datasets and improves in the rest). All the subsequent experiments in the present paper use this validation strategy to choose the prior and posterior. 
\tablename{ \ref{tab:validation}} also includes a comparison of the posterior and prior in terms of the stochastic 01 error and the deterministic 01 error of the mean of the distribution. The results show a consistent improvement of the stochastic error of the posterior over the stochastic error of the prior, while often the error of the mean of the distribution is kept similar or slightly worsened.

\subsection{E3: Percentage of data devoted to prior learning}

Table \ref{tab:percprior} 
shows the results obtained when using different percentages of the dataset to learn the prior by ERM (the rest of the dataset being used to certify the predictor). The first relevant conclusion is that data-free priors (represented in the table as 0\%) are far from competitive, thus justifying that the tightness of the risk certificates are indeed due to the use of data-dependent priors. The table also shows that the amount of data for the prior that leads to the best risk certificates is dataset dependent, with 50\% and 75\% approaching the optimal trade-off. 
Interestingly, using more data for the prior (e.g. 90\%) improves the performance of the predictor but often harms the risk certificates, as the number of data points used to evaluate the bound decreases. This is in line with the findings of \cite{dziugaite2021role}.
We note that the best trade-off on the amount of data for prior building and evaluation of the risk certificate may depend on the size of the dataset. For example, in the case of electricity, which is the second largest dataset, using 90\% of the data for building the prior does not harm the risk certificate considerably (0.2\%), as we will still be using a reasonable amount to evaluate the bound. It is also worth noting the tightness of the risk certificate in this case (2.2\% risk certificate vs 2.1\% stochastic test set 01 error). The same is applicable to MNIST, where the risk certificates improve when using 90\% of the data for prior building (since we will still be using 6000 examples for computing the risk certificate). However, in the case of Bioresponse and Spambase (smallest datasets) the risk certificates worsen by 6.4\% and 4.6\% respectively when comparing 50\% of the data for prior building to 90\%. 
The table also includes the results of the ERM baseline.
For ERM, we used the same range for optimising the learning rate, momentum and dropout rate.  However, given that in this case we do not have a risk certificate we need to set aside some data for validation and hyper-parameter tuning. In all cases we set 5\% of the data for validation. 
Note that this predictor is deterministic, which usually shows slightly better performance than the stochastic counterpart (as also discussed by \cite{dziugaite2018data,perez2020}). The results on these datasets show a slightly larger gap between PNNs and standard neural networks learnt by ERM when compared to the results on MNIST and CIFAR-10 in \cite{perez2020}. Nonetheless, the risk certificates are tight.

\subsection{E4: Differently parametrised architectures}

The experiments in \cite{perez2020} also showed that the KL term 
decreased with the depth of the architecture in CIFAR-10. This is a noteworthy observation and possibly non intuitive initially, as there are many more parameters in the computation of the KL for deep architectures and one would expect the KL term to increase significantly with the number of parameters. However, the authors conjectured that this may be because in a higher-dimensional weight space, the weight updates have smaller euclidean norms (see the discussion in \cite{Zhang2019AreAL}), hence the smaller KL.
We further validate with our experiments that the KL term does not increase with the number of parameters in the architecture. The results are shown in Table \ref{tab:archs}, where we test multiple fully connected architectures architectures with different amount of hidden layers and neurons. 
Our results demonstrate that highly over parameterised models show a KL term of similar factor than those architectures with less parameters.
This  could be because over parameterised models have optimization properties qualitatively different from those that are under parameterised \cite{Belkin15849} or that over parameterised priors are so efficient that posteriors do not need to travel far. Either way, we believe this to be a promising result, demonstrating that there is potential for tight risk certificates even for highly over parameterised architectures. 

\subsection{E5: Training objectives for learning the prior}

The results with different training objectives to learn the prior distribution are gathered in Table \ref{tab:objects}. The first conclusion is that regularisation of the prior using dropout improves the performance of $f_\texttt{erm}$ consistently. However, mixup regularisation seems to worsen the risk certificates and the stochastic error, at least for tabular datasets and FCNs. Note that this objective has an additional hyper-parameter for which we do the same grid sweep as suggested in \cite{mixup} for MNIST. 
The Bayesian inspired learning objective $f_\texttt{bbb}$, which learns the full prior distribution, improves the results over $f_\texttt{erm}$, generally both in terms of the risk certificate and the stochastic 01 error. In this sense, uninformed priors centered at zero, as suggested by \cite{blundell2015weight}, seem to perform better than random uninformed priors. 
Note that $f_{\mathrm{bbb}}$ requires an additional sweep over a KL trade-off coefficient (which attenuates the effect of the KL in the training objective) 
which was done with values in $[1\mathrm{e}^{-6}, 1\mathrm{e}^{-5}, \ldots, 1\mathrm{e}^{-2}]$, see \cite{blundell2015weight}.
The results using the original $f_\texttt{quad}$ objective for learning the prior led to vacuous bounds, so we experimented with a version of this objective that also attenuates the KL (see \cite{perez2020}). The results in Table \ref{tab:objects} include the results of $f_{\mathrm{quad}}$ with the same KL trade-off coefficient sweep as in $f_{\mathrm{bbb}}$. In this case, the risk certificates are not vacuous although not as tight as when learning the prior with $f_{\mathrm{bbb}}$. We conjecture that $f_\texttt{quad}$ still puts much more emphasis on the KL term than $f_{\mathrm{bbb}}$, which in this case measures the divergence to an uninformed prior, which restricts the learning significantly. 
Given that mixup regularisation was originally proposed for vision datasets, convolutional architectures and data starvation scenarios \cite{kimura2021mixup}, we set to experiment with its use in MNIST with a CNN with $4$ layers  where we remove training data at random, reducing the training set to 25\% of the original. The results are in Table \ref{tab:mixup}: note that dropout also helps in this case to learn better PAC-Bayes priors. Additionally, $f_\texttt{mixup}$, which can be seen as a type of data augmentation, leads to very promising results in this case. Specially, it can be seen that $f_\texttt{mixup}$ is data efficient, in the sense that we may need less data to build the prior to achieve tight risk certificates and accurate classifiers (compare the results of $f_\texttt{mixup}$ with 10\% of the data to build the prior to $f_\texttt{erm}$ with 50\% of the data for the prior).

\begin{table}[t]
\centering
\scriptsize
\setlength{\tabcolsep}{2pt}
 \begin{tabular}{|c | c | cc|} 
 \hline
Perc. of data for prior & Prior objective	&		R. Cert. 	&	 S. 01 err. \\ \hline	
\multirow{3}{*}{10\%} & $f_{\texttt{erm}}$	&$	0.110	$&$	0.074$ 
\\	
& $f_{\texttt{erm}}$ + dropout	&$	0.106	$&$	0.067	$
\\
 & $f_{\texttt{mixup}}$	&$	0.065	$&$	0.041	$
\\ \hline
\multirow{3}{*}{25\%} & $f_{\texttt{erm}}$	&$	0.089	$&$	0.056$ 
\\	
& $f_{\texttt{erm}}$ + dropout	&$	0.044	$&$	0.029	$
\\
 & $f_{\texttt{mixup}}$	&$	0.041	$&$	0.025	$
\\ \hline

\multirow{3}{*}{50\%} & $f_{\texttt{erm}}$	&$	0.061	$&$	0.042$ 
\\	
& $f_{\texttt{erm}}$ + dropout	&$	0.026	$&$	0.020	$
\\
 & $f_{\texttt{mixup}}$	&$	0.025	$&$	0.019	$
\\ \hline
\end{tabular} 
\caption{Results with deterministic training objectives to learn the prior on MNIST in a data starvation experiment (reducing the training set in MNIST to 25\% of the original). 
}
\label{tab:mixup}
\vspace*{-0mm}
\end{table}

\begin{figure}[t]
\begin{center}
    \centerline{\includegraphics[width=0.5\textwidth]{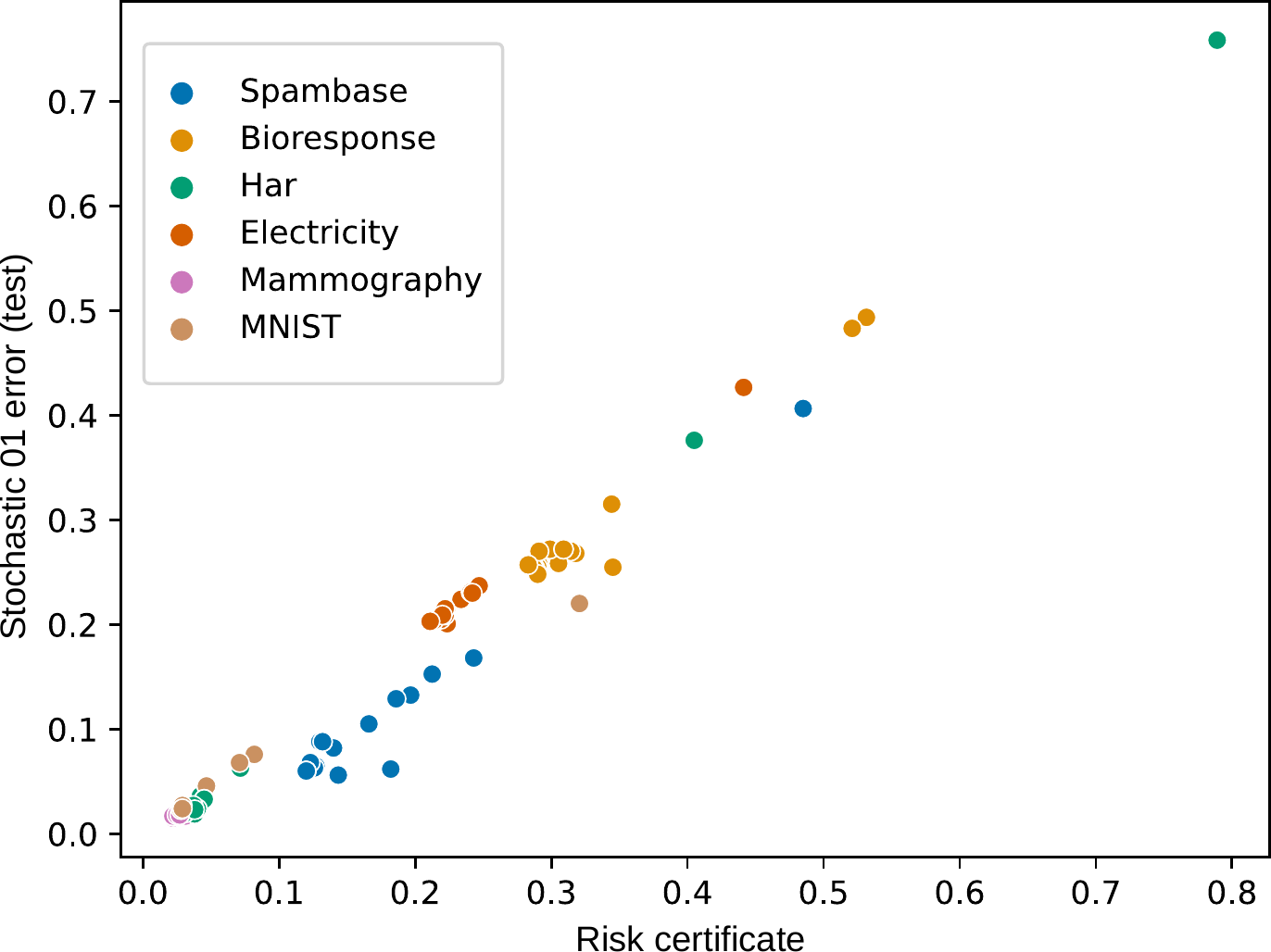}}
    \caption{Consistency of the tightness of the risk certificates across datasets. }
    \label{fig:consistency}
\end{center}
\vspace*{-7mm}
\end{figure}


\paragraph{Tightness across datasets.}
Risk certificates are consistently tight across multiple datasets, as shown in \figurename{~\ref{fig:consistency}}, where we plot all results reported in this paper together. 








	


	






		     


\section{Conclusion}

This work empirically studies learning  PAC-Bayes priors from data. Our results show that data-dependent priors lead to consistently tight risk certificates in $6$ datasets for probabilistic neural classifiers and over parameterised networks, setting a stepping stone towards achieving self-certified learning. We compare a wide range of training objectives for learning the prior distribution, and show that regularisation of the prior is important and that Bayesian inspired learning objectives hold potential for learning appropriate priors, in this case learning the full prior distribution, as opposed to only the mean. Our results also demonstrate that data augmentation may be desirable during prior learning.

\section{Acknowledgments}

We gratefully acknowledge support and funding from the U.S. Army Research Laboratory and the U. S. Army Research Office, and by the U.K. Ministry of Defence and the U.K. Engineering and Physical Sciences Research Council (EPSRC) under grant number EP/R013616/1.

We warmly acknowledge the Department of Computer Science and the AI Centre at University College London, and the Department of Electric and Electronic Engineering at the University of Surrey, for providing friendly and stimulating work environments.

Omar Rivasplata gratefully acknowledges sponsorship from DeepMind for conducting research studies in machine learning at University College London.



\end{document}